\documentclass{article} %
\usepackage{iclr2027_conference,times}

\usepackage{amsmath,amsfonts,bm}

\usepackage{float}
\usepackage{placeins}
\usepackage{hyperref}
\usepackage{url}
\usepackage{booktabs}
\usepackage{multirow}
\usepackage{makecell}
\usepackage{pifont}
\usepackage{amssymb}
\usepackage{xcolor}
\usepackage{colortbl}
\definecolor{ORAVimage}{HTML}{0DC8F2}
\definecolor{ORAVvideo}{HTML}{FF7890}
\definecolor{ORAVaudio}{HTML}{FFCA4E}
\usepackage{microtype}
\usepackage{xurl}
\usepackage{graphicx}
\usepackage{tabularx}
\usepackage{enumitem}
\usepackage{etoolbox}
\makeatletter
\patchcmd{\@maketitle}
  {\hsize\textwidth}
  {\hsize\textwidth\centering}
  {}
  {\errmessage{title centering patch failed}}

\patchcmd{\@maketitle}
  {\begin{tabular}[t]{l}\bf\rule{\z@}{24pt}\@author\end{tabular}}
  {\begin{tabular}[t]{c}\bf\rule{\z@}{16pt}\@author\end{tabular}}
  {}
  {\errmessage{author centering patch failed}}

\patchcmd{\@maketitle}
  {\vskip 0.3in minus 0.1in}
  {\vskip 6pt}
  {}
  {\errmessage{author spacing patch failed}}
\makeatother

\title{ORAV: Benchmarking Audio-Video Generation \\ from Multimodal Contexts}

\author{Jiacheng Hua$^{1,2}$ \quad Xiaokun Feng$^{2}$ \quad Jiaqi Hua$^{1}$ \quad Chang Liu$^{1}$ \quad Yilin Wang$^{1}$ \\
\bf Biao Wang$^{2\dagger}$ \quad Miao Liu$^{1\ddagger}$ \\
$^{1}$College of AI, Tsinghua University \quad $^{2}$Tencent Hy\\
\texttt{hjc21@mails.tsinghua.edu.cn} \quad \texttt{miaoliu@mail.tsinghua.edu.cn}
}

\newcommand{\cmark}{\textcolor{green!45!black}{\ding{51}}}

\newcommand{\numProviders}{5}
\newcommand{\numTasks}{380}
\newcommand{\numBalanced}{350}
\newcommand{\topWinRate}{68.4\%}
\newcommand{\bottomWinRate}{29.9\%}
\newcommand{\numVariants}{12}
\newcommand{\vetoShare}{18.2\%}
\newcommand{\substitutionShare}{95\%}
\newcommand{\vetoDecided}{27\%}
\newcommand{\vetoDecidedVideo}{32\%}
\newcommand{\vetoWorstVideo}{56\%}
\newcommand{\vetoWorstNoVideo}{12\%}
\newcommand{\replayWan}{9/78}

\newcommand{\humanAgreementSelected}{86.08\%}
\newcommand{\humanReviewSampleShare}{10\%}

\newcommand{\speechOnlyMini}{77.8\%}
\newcommand{\speechOnlySeed}{58.3\%}
\newcommand{\speechMotionSeed}{73.4\%}
\newcommand{\speechMotionMini}{53.6\%}

\newcommand{\qualityWideGap}{0.266}
\newcommand{\qualityWideCI}{[0.205, 0.328]}

\renewenvironment{abstract}{%
  \vskip 2pt\centerline{\large\sc Abstract}\vspace{0.5ex}%
  \begin{list}{}{\leftmargin 0.25in\rightmargin 0.25in\topsep 4pt\partopsep 0pt}%
  \item\relax
}{\par\end{list}\vskip 1ex}
\iclrfinalcopy %
\newcommand{\arxivReleaseStatement}{Code and benchmark data will be released upon approval.}
\begin{document}

\maketitle
\begin{NoHyper}{\renewcommand{\thefootnote}{\ensuremath{\dagger}}%
\footnotetext{Project lead.\quad$^{\ddagger}$Corresponding author.}}\end{NoHyper}
\fancypagestyle{arxivfirst}{%
  \fancyhf{}%
  \fancyhead[L]{\raisebox{-2pt}[0pt][0pt]{\includegraphics[height=16pt]{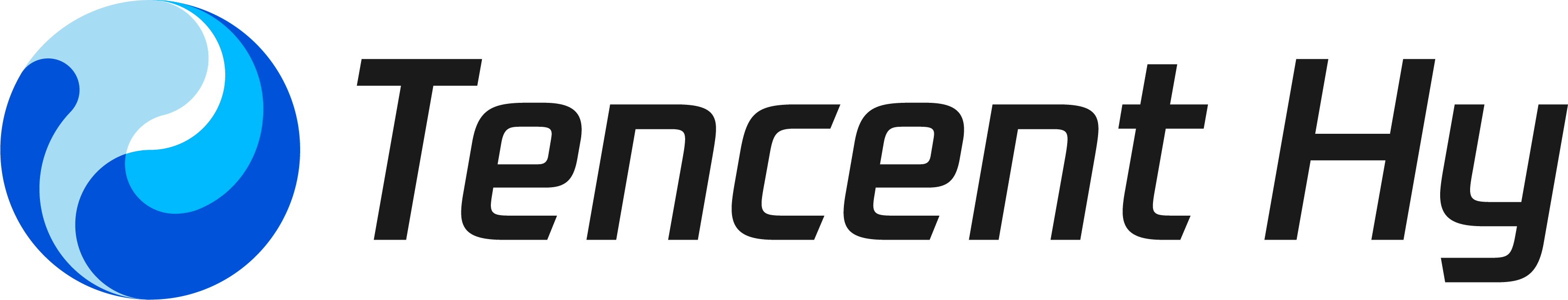}}%
    \hspace{8pt}\raisebox{3pt}[0pt][0pt]{\textcolor{black!55}{\large$\times$}}\hspace{8pt}\raisebox{-2pt}[0pt][0pt]{\includegraphics[height=16pt]{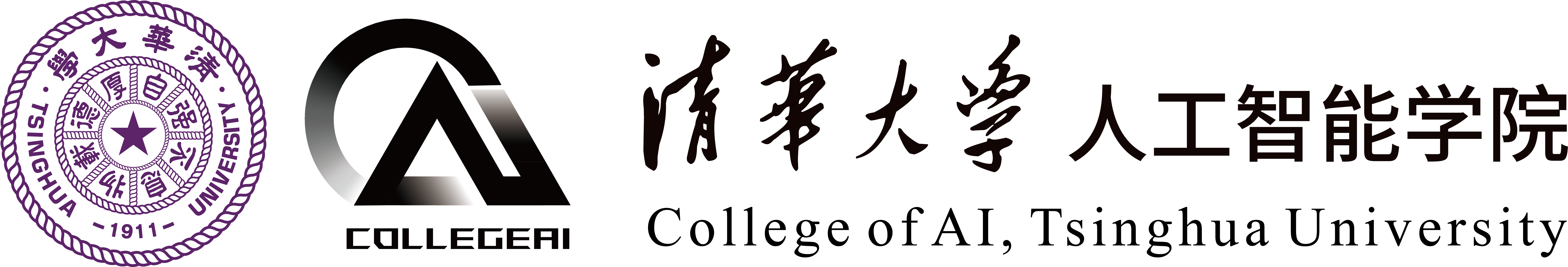}}}%
  \fancyfoot[C]{\thepage}%
  \renewcommand{\headrulewidth}{0.4pt}}
\thispagestyle{arxivfirst}
\lhead{}
\renewcommand{\headrulewidth}{0pt}

\begin{abstract}

Audio-video generation using heterogeneous multimodal references has emerged as a new challenge, requiring both compositional control over generation and grounded understanding of multimodal context. In this paper, we introduce \textbf{ORAV Bench} for \textbf{O}mni \textbf{R}eference \textbf{A}udio-\textbf{V}ideo Generation, comprising \numTasks{} task instances with 2--10 references, 9 semantic roles, and 30 role compositions. Instructions specify the relationships among references; the media supply the identities, dynamics, and audio characteristics to be realized. To evaluate these open-ended outputs, we develop a reference-aware pairwise protocol that prepares visual and auditory evidence, compares the intended contribution of each reference, and checks the overall verdict in both presentation orders. On held-out instances, it achieves \humanAgreementSelected{} effective agreement with human judgments. Across \numProviders{} frontier systems, overall rankings conceal distinct strengths across reference compositions. A recurring failure is to reproduce unintended source content in place of the requested result, despite closely resembling a reference. Reproducible pointwise diagnostics of quality, reference affinity, and speech reveal distinct dimensions of model behavior. ORAV thus offers a benchmark for tracking progress toward controllable, compositional, and reference-faithful audio-video generation.

\end{abstract}

\suppressfloats
\section{Introduction}

\begin{figure}[t]
\centering
\includegraphics[width=\linewidth]{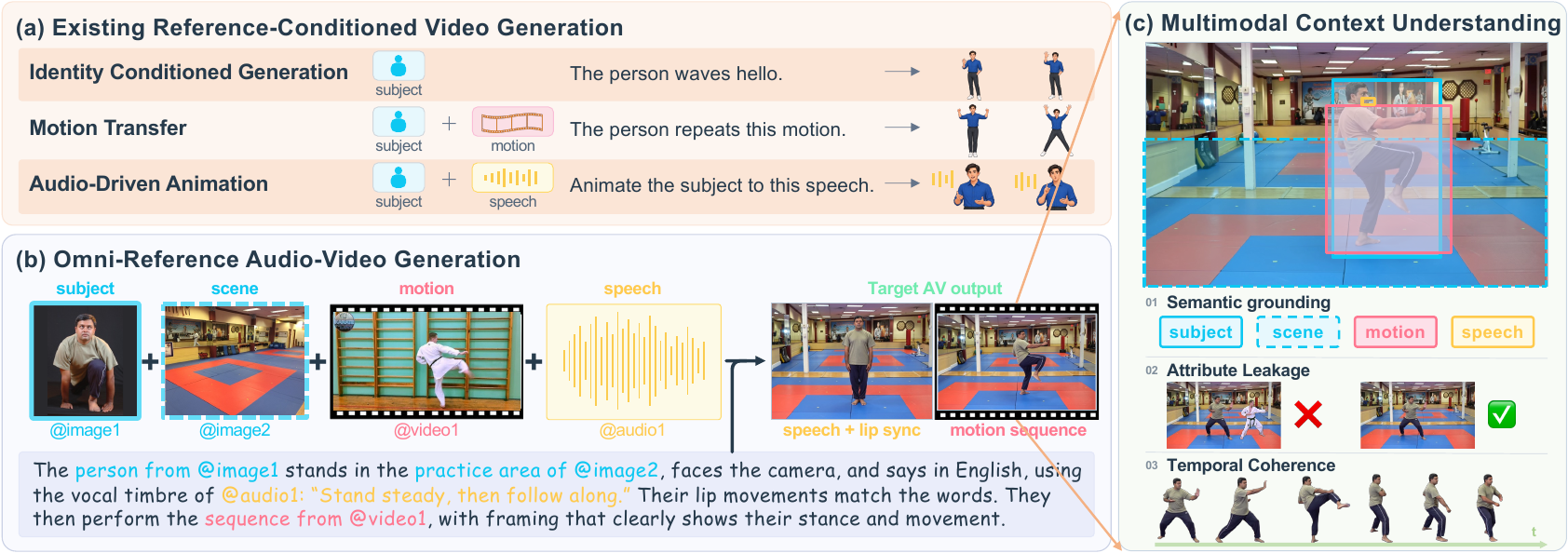}
\vspace{-1.5em}
\caption{
\emph{Omni-reference audio-video generation requires selectively composing information from heterogeneous references.}
(a) Representative reference-conditioned task types include identity-conditioned generation, motion transfer, and audio-driven animation.
(b) An omni-reference task instance combines subject and scene images, a motion video, and a speech reference under a textual instruction. The target subject speaks the specified utterance using the referenced vocal timbre, then performs the referenced motion in the designated scene. 
(c) Successful generation preserves assigned properties, prevents attribute leakage, and coherently combines speech and motion.
}
\label{fig:teaser}
\end{figure}

Generative models have enabled realistic synthesis of images, videos, and audio-video content \citep{rombach2022ldm,chen2026skyreelsv4}. A growing focus is to make this process more precisely and flexibly controllable through user-provided references. Such references can convey a person's appearance, a movement, or a voice more directly than words alone \citep{jiang2023videobooth,zhao2023motiondirector,lin2025omnihuman}, as illustrated in Fig.~\ref{fig:teaser}a. Frontier systems now accept heterogeneous image, video, and audio references together \citep{wan2025wanopenadvancedlargescale,klingteam2025klingomnitechnicalreport,seedance2026seedance20advancingvideo,chen2026skyreelsv4,vorchomni2026,minimax2026h3context}, enabling the omni-reference audio-video generation setting in Fig.~\ref{fig:teaser}b. Using these inputs together requires \emph{multimodal context understanding and generation}: understanding how textual instructions assign contributions to the references and composing them into the requested audio-video result.

Beyond its practical value for fine-grained content creation, omni-conditioning raises a more fundamental scientific question about the compositional capabilities of generative models: whether existing generative methods can maintain controllability when composing heterogeneous, cross-modal conditions. Studying this capability requires a systematic benchmark that pairs multimodal references with textual instructions, specifying how individual references should be composed into semantically plausible and contextually grounded audio-video outputs. Existing audio-video benchmarks emphasize perceptual quality \citep{huang2024vbench,liu2024evalcrafter}, fidelity to individual references \citep{yuan2025opensvnexus}, and audio-video coherence \citep{cao2026tavcompass,hua2026vabench}, offering complementary perspectives on how audio-video outputs are rendered and controlled, but lack systematic evaluation of compositional controllability under heterogeneous, cross-modal references, leaving omni-reference audio-video generation insufficiently assessed.

To bridge this gap, we introduce \textbf{ORAV Bench}, a benchmark for \textbf{O}mni \textbf{R}eference \textbf{A}udio-\textbf{V}ideo Generation.\ifdefined\arxivReleaseStatement\footnote{\arxivReleaseStatement}\fi{} ORAV covers diverse multimodal reference conditions, and organizes them into a hierarchical taxonomy that supports both coarse-grained reporting and fine-grained diagnostic analysis. Each instance specifies a composition contract that defines which reference should control which output factor, enabling ORAV to probe whether models can selectively extract relevant information, bind it to the intended role, and realize it in the generated audio-video output.

Evaluating ORAV requires understanding the intended use of a reference, as illustrated in Fig.~\ref{fig:teaser}c. An output may inherit the performer and setting of a motion source, or preserve a voice by replaying words the task asks it to replace. Such outcomes reveal a gap between resemblance to source material and fulfillment of the requested composition. We develop a reference-aware pairwise evaluator that interprets visual and auditory evidence through each reference's semantic contract. It compares the assigned contributions and their binding, then judges how their differences affect the complete event. This connects overall preferences to identifiable successes and failures in using multimodal context.

Across \numProviders{} frontier systems, overall rankings conceal complementary strengths on different reference compositions. The evaluation also reveals a recurring failure: copying a reference instead of using it as instructed, such as reproducing the original performers and scene of a motion video in place of the requested ones. Quality and source-affinity diagnostics further separate well-rendered or reference-like outputs from successful task fulfillment. These findings establish selective composition of multimodal context as a distinct evaluation target.

Our work therefore makes three main contributions:
\vspace{-0.5em}
\begin{itemize}[leftmargin=*,itemsep=1pt,topsep=2pt]
    \item We introduce ORAV Bench for omni reference audio-video generation, making multimodal context understanding and generation testable through diverse reference roles and compositions.

\item We develop a reference-aware pairwise evaluation protocol that connects task-level preferences to visual and auditory evidence, and validate it through held-out human evaluation.

    \item We evaluate audio-video systems on ORAV and analyze their capabilities and failure modes, revealing challenges in composing heterogeneous references. Reproducible pointwise metrics provide a diagnostic interface for quality, reference affinity, and speech following.
\end{itemize}

\section{Related Work}
\subsection{Reference Conditioned Video Generation}

Reference-conditioned generation uses images, videos, or audio to guide synthesis. Prior work has explored subject preservation and identity binding \citep{jiang2023videobooth,wei2023dreamvideo,deng2025magref}, motion transfer and trajectory control \citep{zhao2023motiondirector,ren2024customizeavideo,wei2024dreamvideo2}, and audio-driven human animation \citep{lin2025omnihuman}. Recent systems support multimodal generation and task orchestration \citep{jiang2025vace,pan2026omniweaving,vorchomni2026}. These advances broaden the conditioning interface, but input modality alone does not specify a reference's intended contribution. ORAV Bench evaluates how models extract and compose information from heterogeneous references under text-specified factor--target contracts.

\begin{table}[t]
  \centering
  \caption{\emph{Comparison with related video and audio--video generation benchmarks.}
  Reference modalities and semantic coverage are pooled across task instances; joint composition reports the
  largest set of semantic families required together from distinct references in a single instance.}
  \label{tab:benchmark_comparison}
  \vspace{2pt}

  \begingroup
  \small
  \setlength{\tabcolsep}{3pt}
  \renewcommand{\arraystretch}{1.12}
  \begin{tabular*}{\linewidth}{@{\extracolsep{\fill}}lcccccc@{}}
    \toprule
    \multirow{2}{*}{\textbf{Benchmark}} &
    \multicolumn{2}{c}{\textbf{Generation}} &
    \multicolumn{3}{c}{\textbf{Reference-derived factors}} &
    \multirow{2}{*}{\makecell{\textbf{Joint}\\\textbf{composition}}} \\
    \cmidrule(lr){2-3}\cmidrule(lr){4-6}
    & Output & References & Content & Dynamics & Audio & \\
    \midrule
    T2AV-Compass~{\scriptsize\citeyearpar{cao2026tavcompass}}
      & AV & -- & -- & -- & -- & -- \\
    VABench~{\scriptsize\citeyearpar{hua2026vabench}}
      & AV & I & \cmark & -- & -- & -- \\
    UI2V-Bench~{\scriptsize\citeyearpar{zhang2025ui2vbench}}
      & V & I & \cmark & -- & -- & -- \\
    LongAV-Compass~{\scriptsize\citeyearpar{liu2026longavcompass}}
      & AV & I, V & \cmark & -- & -- & -- \\
    OpenS2V-Eval~{\scriptsize\citeyearpar{yuan2025opensvnexus}}
      & V & I & \cmark & -- & -- & C \\
    UniVBench$^{\ddagger}$~{\scriptsize\citeyearpar{wei2026univbench}}
      & V & I & \cmark & -- & -- & C \\
    MotionBench~{\scriptsize\citeyearpar{ma2026effivmt}}
      & V & V & -- & \cmark & -- & -- \\
    MSAVBench~{\scriptsize\citeyearpar{wei2026msavbench}}
      & AV & I, A & \cmark & -- & \cmark & C\,+\,A \\
    MultiRef-Compass~{\scriptsize\citeyearpar{zhang2026multirefcompass}}
      & AV & I, V$^{\dagger}$, A$^{\dagger}$ & \cmark & -- & \cmark$^{\dagger}$ & (C\,+\,A)$^{\dagger}$ \\
    \midrule
    \textbf{ORAV Bench (ours)}
      & AV & I, V, A & \cmark & \cmark & \cmark & \textbf{C\,+\,D\,+\,A} \\
    \bottomrule
  \end{tabular*}
  \endgroup

  \smallskip
  \begin{minipage}{\linewidth}
  \footnotesize\raggedright
  \textit{Notation.} I/V/A: image/video/audio references; V/AV: video/audio--video output.
  Reference-derived factors: C, content/appearance; D, motion/camera movement/visual effects;
  A, audio attributes. C alone denotes within-family composition.
  \cmark/--: included/not included in the listed setting;
  $^{\dagger}$: extended track; $^{\ddagger}$: T2V/R2V only.
  \end{minipage}
\end{table}

\subsection{Video and Audio-Video Generation Evaluation}
Video-generation benchmarks assess perceptual quality, temporal consistency, and semantic alignment \citep{huang2024vbench,liu2024evalcrafter}. Recent work extends evaluation to audio quality and audio-video coherence \citep{cao2026tavcompass,hua2026vabench}, minute-scale and multi-shot generation \citep{liu2026longavcompass,wei2026msavbench}, and reference-based subject preservation \citep{yuan2025opensvnexus}. Most closely related, MultiRef-Compass evaluates multi-reference fidelity, binding, and audio-video consistency \citep{zhang2026multirefcompass}; its visual references primarily specify subject, object, and scene appearance. ORAV Bench asks a complementary, explicitly factorized question: whether generators can selectively extract, bind, and compose information from heterogeneous references across modalities, entities, and time. Its coverage includes reference-supplied motion, camera movement, and visual effects alongside appearance and audio semantics, as summarized in Tab.~\ref{tab:benchmark_comparison}. The evaluation assesses how well each generated output jointly satisfies the prompt-specified factor–target contracts.

\subsection{Pairwise Evaluation and Model-Based Judging}
Arena-style evaluation derives model rankings from blind comparisons between outputs generated for the same input \citep{jiang2024genaiarena}. Prior work explores human-aligned automated pairwise judging \citep{li2026genarena} and trains video reward models using multidimensional human preferences \citep{liu2025videohumanfeedback}. Agent-based and multimodal evaluators incorporate structured prompts, temporal tools, automatic metrics, and cross-model rejudging \citep{yang2025videogeneval,zhang2026multirefcompass}, while model judges remain susceptible to order and self-enhancement biases \citep{zheng2023judging}. Building on these evaluation practices, ORAV organizes comparisons around the intended contribution of each heterogeneous reference and its target in the output. Factor-level diagnostics complement the rankings by identifying failures of transfer, binding, and selection of the intended source content.

\section{Benchmark Design}
\label{sec:design}

\begin{figure}[t]
\centering
\includegraphics[width=\linewidth]{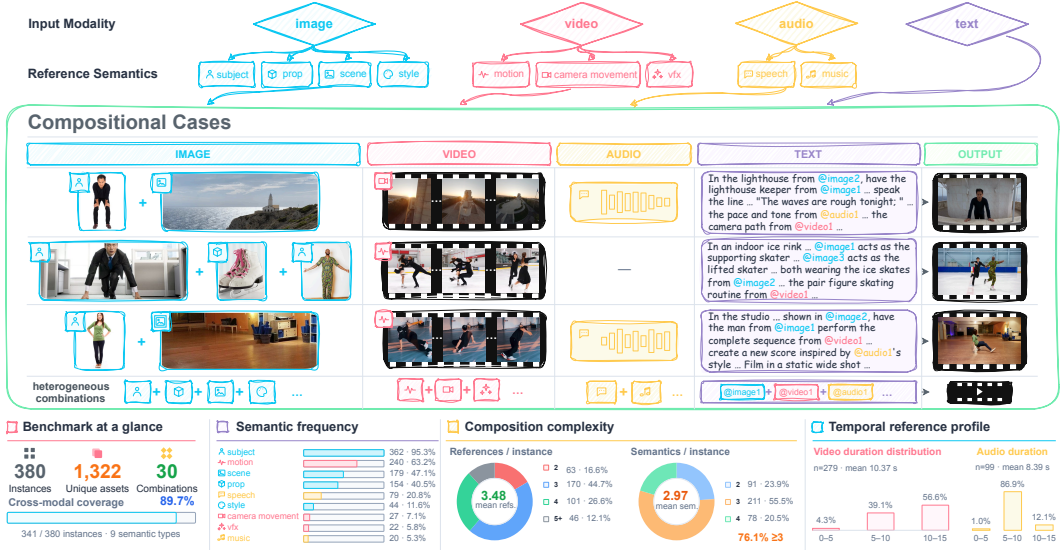}
\vspace{-1.5em}
\caption{
\emph{Overview of the benchmark.}
Top: Nine semantic roles describe the intended contributions of image, video, and audio references.
Middle: Representative task instances combine heterogeneous references with a global textual instruction, which specifies what to extract from each reference and how to bind and compose the selected information in the generated audio-video output.
Bottom: Benchmark statistics summarize dataset scale, semantic-role frequencies, composition complexity, and video/audio reference-duration distributions.
}
\label{fig:overview}
\end{figure}

Our benchmark studies \emph{multimodal context understanding and generation} through omni-reference audio-video synthesis. Each instance pairs multimodal references with an instruction specifying their contributions and relationships. Fig.~\ref{fig:overview} summarizes ORAV's semantic roles, compositions, and scale.

\subsection{Task Formulation}
\label{sec:design:formulation}

In \emph{omni-reference audio-video generation}, a model synthesizes video and audio from a textual instruction and a collection of references drawn from image, video, and audio modalities. Let
\[
    \mathcal{R}
    =
    \mathcal{R}^{I}
    \cup
    \mathcal{R}^{V}
    \cup
    \mathcal{R}^{A}
    =
    \{r_1,\ldots,r_J\},
\]
where $\mathcal{R}^{I}$, $\mathcal{R}^{V}$, and $\mathcal{R}^{A}$ denote the image, video, and audio reference sets, respectively, and $J$ is the total number of references. Given instruction $T$ and references $\mathcal{R}$, the model $G_{\theta}$ generates
\[
    \hat{Y}
    =
    G_{\theta}(T,\mathcal{R})
    =
    (\hat{Y}^{v},\hat{Y}^{a}),
\]
where $\hat{Y}^{v}$ and $\hat{Y}^{a}$ are the generated visual and audio streams. The output should be perceptually natural and temporally coherent across audio and video, while faithfully realizing the semantic requirements and cross-reference relationships specified by $T$ and $\mathcal{R}$.

Crucially, references are not intended to be reproduced in their entirety. Each reference has an intended \emph{semantic role} that specifies which information it should contribute. For example, a video may provide motion without providing actor identity, while an audio clip may provide speaker timbre without providing its original lexical content. The model must therefore determine \emph{what} to extract from each reference, \emph{where} to apply it, and \emph{what} source information to ignore.

We make this intended use explicit through a \emph{semantic reference contract}. For reference $r_j$, the instruction assigns a role $s_j=\rho_T(r_j)$ from the nine-role vocabulary $\mathcal S$. Its contract is
\[
    c_j=\bigl(r_j,\ s_j,\ b_j,\ \mathcal P_j,\ \mathcal N_j\bigr),
    \qquad s_j\in\mathcal S.
\]
Here $r_j$ identifies the source, $s_j$ specifies its semantic use, and $b_j$ identifies the output target, such as a person, event, or audio track. The requirements $\mathcal P_j$ describe what should be realized from the source, including task-permitted variation in viewpoint or timing. The exclusions $\mathcal N_j$ describe source content whose transfer would violate the instruction or another reference's assignment; this set can be empty. These components formalize the task semantics; the judge interprets them directly from the English instruction and reference media.

As illustrated in Fig.~\ref{fig:teaser}b, the motion contract preserves the demonstrated movement while excluding the original performer and background, whereas the speech contract preserves vocal character while enforcing the instructed utterance. These contracts disentangle transferable attributes from conflicting source content, with modality providing the evidence and the contract specifying the required correspondence. The complete task combines the contracts $\mathcal C(T,\mathcal R)=\{c_j\}_{j=1}^{J}$,
with instruction-level binding and temporal relationships among their targets. An output may show the requested person while someone else performs the reference motion. Both sources are recognizable, but the requested subject--action binding is missing.

\subsection{Benchmark Construction}
\label{sec:design:construction}
\noindent\textbf{Semantic role and composition design}.
Community use cases and expert discussions inform nine semantic roles, organized by the intended contribution of each reference. Image references specify content and appearance through \texttt{subject}, \texttt{prop}, \texttt{scene}, and \texttt{style}. Video references provide dynamic visual cues through \texttt{motion}, \texttt{camera\_movement}, and \texttt{vfx} (visual effects). Audio references provide \texttt{speech} and \texttt{music}.

To test how reference contributions work together, we group task instances by their \emph{composition signature}: the set of roles assigned to the references, irrespective of multiplicity. For example, separate subject images and a group photograph can yield the same signature but require different subject bindings. We select compositions from community examples and expert analysis of events that naturally combine complementary reference contributions. The resulting signatures support comparison across task contexts, with coverage across signatures rather than equal instance counts. %

\noindent\textbf{Task instantiation and binding}.
Task instantiation requires both observable source attributes and a feasible joint event. We combine task-driven retrieval with reference-driven design: proposed scenarios guide media search, and suitable source material can suggest new instances. Candidates are drawn from existing media collections and supplementary public sources, then screened for clarity, suitability, relevance, and observability of the intended semantic attribute. Video references undergo human inspection, while audio screening combines model-assisted inspection with signal-level checks. During task assembly, we check cross-reference compatibility: scenes must accommodate the requested action and required facilities, props must support the intended interactions, and all necessary participants must be specified. A motion reference contributes the demonstrated action; the assigned subject and scene may come from different sources.

Instructions establish how the selected sources should contribute to one output. Each reference receives a unique handle, such as \texttt{@image1}, \texttt{@video1}, or \texttt{@audio1}. The text specifies the objective, role assignments, and relationships; the references retain the detailed perceptual information needed to realize them. Spatial cues such as ``the person on the left'' are anchored to a specified reference image or video frame. Instructions for multi-person actions explicitly bind target subjects to the corresponding performers or actions. Instances without a scene reference receive environmental context in text. Speech instructions specify the words to be spoken and the intended on-screen speaker. For instances coupling music and motion, instructions identify which reference defines the timing and what temporal adjustments are permitted.

\noindent\textbf{Quality control and benchmark scale}.
Model-assisted checks and repeated human review verify that each instance defines an observable, consistent contract through reference-role alignment, clear bindings, complete participants and facilities, and instruction--media consistency. Review feedback drives revisions to references, instructions, and bindings; modified instances are flagged for renewed review. Human experts make all final inclusion decisions. For each instance, the total duration of video references and the total duration of audio references are each capped at $15\,\mathrm{s}$, with excerpts trimmed as needed. These limits are chosen to accommodate the reference-input constraints of the evaluated systems. Packaged instances include media assets, textual instructions, reference-handle mappings, and source and review metadata.

The resulting benchmark contains \numTasks{} task instances spanning 30 composition signatures and using 1,322 unique reference assets. Each instance includes 2--10 references covering 2--4 semantic roles. References are counted as media assets rather than individual subjects depicted within them. Text is part of the conditioning context but is excluded from both counts. Reference assets are not reused across instances, reducing cross-instance correlations and preventing repeated source content from disproportionately influencing aggregate results.

\section{Reference-Aware Pairwise Evaluation}
\suppressfloats[t]
\label{sec:eval}

\begin{figure}[t]
    \centering
    \includegraphics[width=\linewidth]{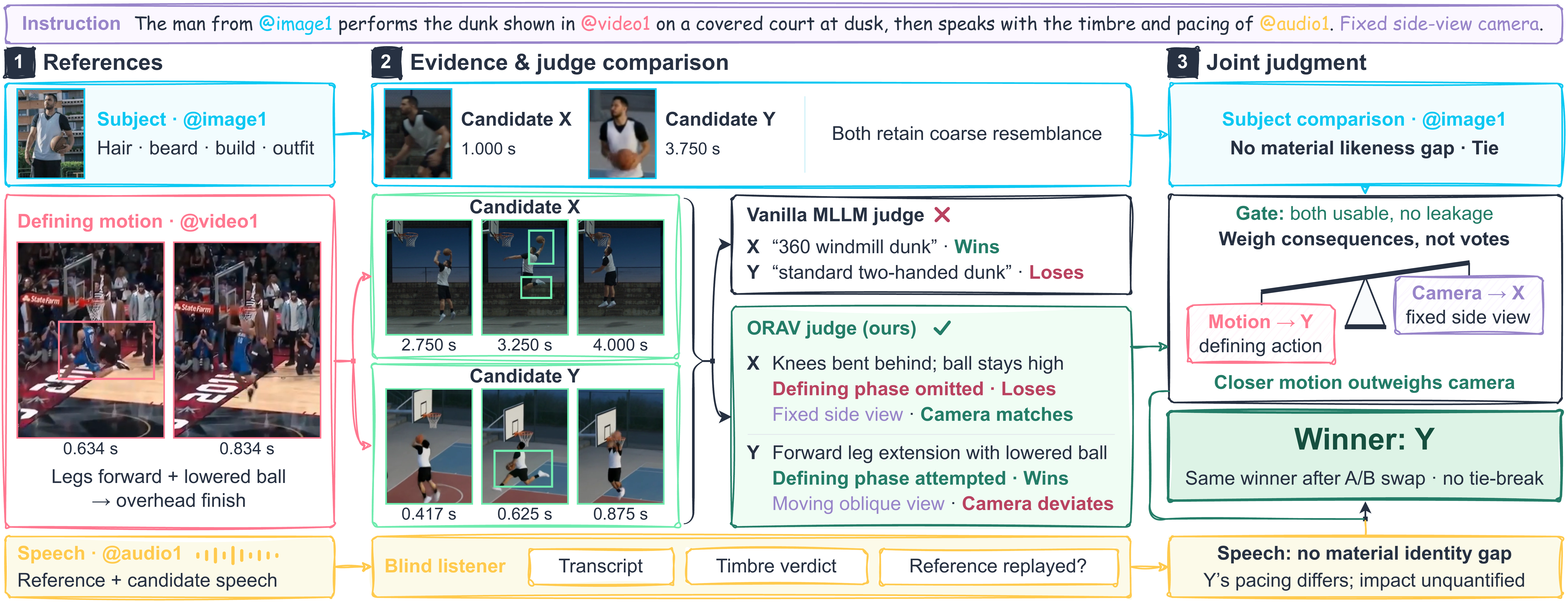}
    \caption{\emph{From reference evidence to a joint judgment.} With subject likeness tied, Y's closer reproduction of forward leg extension with a lowered ball outweighs X's better fixed-camera compliance. The visual case contrasts Gemini-3.1-Pro-Preview on native video with GPT-6-Astra on panels; the speech branch illustrates the integration of auditory evidence.}
    \label{fig:pipeline}
\end{figure}

Successful reference use requires both correspondence with the assigned sources and coherence among their contributions. The evaluator in Fig.~\ref{fig:pipeline} reasons across these two levels: it establishes what each candidate realizes from each reference, then determines how those contributions combine into the requested event. Semantic contracts define the comparison; visual and auditory observations support it; task-level reasoning turns the supported differences into an overall preference.

\subsection{Comparing Reference Use in Context}
\label{sec:eval:contract}\label{sec:eval:why}
Reference fidelity depends on intended use. A motion clip may supply an action but not its performer or setting; a speech recording may supply vocal character for a new utterance. The contracts in Sec.~\ref{sec:design:formulation} formalize this selective correspondence: properties $\mathcal P_j$ should appear at target $b_j$, subject to exclusions $\mathcal N_j$. Reference matching assesses each source's assigned contribution.

Movements, camera paths, and spoken utterances call for different evidence. Comparing two candidates under the same instruction and references lets the judge assess each requirement on its own terms and express the result as a relative preference. For each reference, candidate-specific evidence precedes the preference in the record
\[
    q_j=\bigl(e_j^A,e_j^B,d_j,o_j\bigr),
    \qquad d_j\in\{A,B,\mathrm{tie}\},
\]
where $e_j^A$ and $e_j^B$ describe the observed correspondence, $d_j$ gives the relative preference, and $o_j$ records evidence sufficiency. This record connects each preference to its supporting observations.

Composition requires these contributions to belong to the same instructed event. Binding judgments ask whether the reference-like subject is the one performing the demonstrated action; instruction judgments connect that event to requirements such as camera placement and temporal order. These relationships link the reference-level comparisons to overall task fulfillment. The judge returns the linked judgments and overall preference together in one call per presentation order.

\subsection{Grounding Comparisons in Visual and Auditory Evidence}
\label{sec:eval:grounding}
Motion fidelity depends on how an action unfolds. Timestamped panels spanning the reference and candidate clips expose phases and trajectories that distinguish movements within the same action category. In the dunk of Fig.~\ref{fig:pipeline}, forward leg extension with a lowered ball distinguishes the demonstrated maneuver from a conventional dunk. The judge compares these mechanics and their sequence, citing supporting timestamps and accounting for permitted timing changes. Its evidence identifies realized and altered action phases, providing a concrete basis for the later preference.

Independent listening supplies the corresponding auditory basis. Without video or system identity, the listener records what is said, how the voice resembles its reference, and whether the source recording is replayed. These observations separate successful utterance generation from vocal resemblance and recording reuse. The final judge interprets them against the requested speech and assigned speaker, bringing auditory reference use into the same task comparison as visual reference use.

\subsection{Combining Evidence into a Preference and Ranking}
\label{sec:eval:aggregation}

The overall preference follows three priorities: \emph{usability}, \emph{task fulfillment}, and \emph{general quality}. Usability asks whether the delivered artifact supplies the requested integrated rendition. Preserving a specified scene or making a permitted timing change remains legitimate, as do harmless incidental details. Source content fails usability when its intrusion materially replaces or obstructs the requested rendition; lesser discrepancies remain task-level differences. The boundary follows the instruction and contracts, rather than a universal similarity or duration threshold. When usability does not distinguish the candidates, the judge weighs supported differences by their magnitude and task consequence; general quality breaks a task tie. Each supported difference counts once, with priorities determined by the task rather than a fixed role hierarchy.

In the underlying dunk task, candidate X supplies the requested static side view but performs a conventional dunk, omitting the reference's forward leg extension with a lowered ball. Candidate Y more closely preserves these defining features, although its camera moves. With subject likeness tied at the available resolution, the judge favors Y: preserving more of the defining action contributes more to the requested event than X's camera advantage. Both presentation orders select Y.

Reversing presentation order checks whether the preference follows candidate identity rather than presentation position. Matching overall decisions are retained; order conflicts contribute a tie for ranking. A Davidson--Bradley--Terry model aggregates the pairwise results on \numBalanced{} shared-delivery instances, with bootstrap intervals clustered by instance. Delivery is reported separately, keeping the ranking focused on task performance among delivered outputs.

\subsection{Held-out Human Validation}
\label{sec:eval:meta}

\begin{figure}[t]
    \centering
    \includegraphics[width=\linewidth]{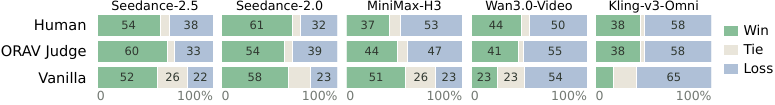}
    \vspace{-1.5em}
    \caption{\emph{Human and automated win--tie--loss profiles.} The three evaluators share 324 candidate pairs from 38 held-out instances, with equal instance weights. Automated ties include AB/BA conflicts.}
    \label{fig:human-wtl}
\end{figure}

Five expert annotators independently performed blind evaluations \citep{bojar-etal-2014-findings} on a random \humanReviewSampleShare{} sample of benchmark instances held out from evaluator development.

We compare against a vanilla Gemini-3.1-Pro-Preview judge on the same 38 held-out instances. Given the instruction, references, and candidate videos, it is explicitly prompted to assess reference fidelity and audio-video quality in both presentation orders. Our evaluator achieves \humanAgreementSelected{} effective human agreement versus 71.82\% for this baseline, averaging instances equally over each evaluator's valid non-tied comparisons.
Fig.~\ref{fig:human-wtl} compares the resulting win--tie--loss profiles. Notably, the vanilla judge tends to make incorrect judgments due to its limited ability to identify subtle differences, whereas our evaluation framework aligns more closely with human annotations.

With the same Astra judge and prepared evidence, ORAV's structured protocol improves paired human agreement by 7.45 percentage points over the simplified prompt in Appx.~\ref{app:prompt}.

\section{Experiments}
\suppressfloats[t]
\label{sec:exp}

We evaluate Seedance-2.5, Seedance-2.0, MiniMax-H3, Wan3.0-Video, and Kling-v3-Omni on \numTasks{} task instances. Generation uses 720p for four systems and 768p for MiniMax-H3, whose available settings are 768p and 2K. Candidate frames are presented at a common width, with task fulfillment prioritized over general quality to limit the influence of sharpness differences. We use GPT-6-Astra for final judgments and Gemini-3.1-Pro-Preview for independent listening.

\subsection{Overall Performance}
\begin{table}[t]
\centering
\definecolor{ORAVoutput}{HTML}{70EBAF}
\caption{\emph{Overall and role-conditioned performance on ORAV.} Systems are ordered by Davidson--Bradley--Terry strength on 350 shared-delivery instances, with 95\% bootstrap intervals over instances. Role columns report expected win rates (\%) for overlapping subsets containing each role, including their co-occurring requirements. Bold marks the highest estimate in each column.}
\label{tab:overall}\label{tab:roles}
\small
\setlength{\tabcolsep}{1.25pt}
\renewcommand{\arraystretch}{1.17}
\begin{tabular*}{\linewidth}{@{\extracolsep{\fill}}r*{12}{c}@{}}
\toprule
\multirow{2}{*}{\#} & \multirow{2}{*}{System} &
\multicolumn{2}{c}{\cellcolor{ORAVoutput!16}Overall} &
\multicolumn{4}{c}{\cellcolor{ORAVimage!16}Image} &
\multicolumn{3}{c}{\cellcolor{ORAVvideo!16}Video} &
\multicolumn{2}{c}{\cellcolor{ORAVaudio!16}Audio} \\
\cmidrule(lr){3-4}\cmidrule(lr){5-8}\cmidrule(lr){9-11}\cmidrule(lr){12-13}
& & \cellcolor{ORAVoutput!8}Strength & \cellcolor{ORAVoutput!8}95\% CI & \cellcolor{ORAVimage!8}Subj. & \cellcolor{ORAVimage!8}Prop & \cellcolor{ORAVimage!8}Scene & \cellcolor{ORAVimage!8}Style & \cellcolor{ORAVvideo!8}Motion & \cellcolor{ORAVvideo!8}Cam. & \cellcolor{ORAVvideo!8}VFX & \cellcolor{ORAVaudio!8}Speech & \cellcolor{ORAVaudio!8}Music \\
\midrule
1 & Seedance-2.5 & $\mathbf{+0.70}$ & {\footnotesize $[+0.56, +0.85]$} & $\mathbf{68.92}$ & $\mathbf{64.07}$ & $\mathbf{68.61}$ & $\mathbf{65.48}$ & $\mathbf{71.01}$ & $65.38$ & $\mathbf{82.63}$ & $68.04$ & $\mathbf{76.56}$ \\
2 & Seedance-2.0 & $+0.48$ & {\footnotesize $[+0.36, +0.59]$} & $62.82$ & $61.01$ & $61.85$ & $63.54$ & $65.70$ & $64.90$ & $71.36$ & $59.95$ & $50.98$ \\
3 & MiniMax-H3 & $+0.15$ & {\footnotesize $[+0.02, +0.28]$} & $53.39$ & $56.40$ & $56.87$ & $57.51$ & $42.37$ & $\mathbf{65.87}$ & $58.28$ & $\mathbf{73.58}$ & $39.85$ \\
4 & Wan3.0-Video & $-0.56$ & {\footnotesize $[-0.70, -0.42]$} & $35.50$ & $38.90$ & $34.71$ & $36.47$ & $30.06$ & $25.00$ & $30.00$ & $40.90$ & $40.42$ \\
5 & Kling-v3-Omni & $-0.77$ & {\footnotesize $[-0.93, -0.63]$} & $29.37$ & $29.63$ & $27.96$ & $27.00$ & $40.86$ & $28.85$ & $7.73$ & $7.53$ & $42.19$ \\
\bottomrule
\end{tabular*}
\end{table}

Tab.~\ref{tab:overall} gives the overall ranking on ORAV. Seedance-2.5 leads, followed by Seedance-2.0 and MiniMax-H3; Wan3.0-Video and Kling-v3-Omni trail. The estimated win rates against a random opponent range from \topWinRate{} to \bottomWinRate{}. The full order is unchanged across \numVariants{} aggregation and data-handling settings. Instance-clustered bootstrap intervals quantify uncertainty, and shared-delivery evaluation fixes the instances across all five systems. The ranking also remains unchanged when missing videos are counted as losses, as detailed in Appx.~\ref{app:ranking}.

On shared-delivery instances, Seedance-2.5 achieves an observed pairwise win rate of 58\% against Seedance-2.0 and 80\% against Kling-v3-Omni, with ties counted as half wins. Its advantage is modest over the runner-up and substantially larger over the lowest-ranked system. The role-conditioned results further reveal which reference contexts favor each system.

\subsection{Context-Dependent Strengths}
Tab.~\ref{tab:roles} refits the overall preference model within instances containing each role. Seedance-2.5 has the highest estimate in seven of nine buckets; MiniMax-H3 leads the speech and camera buckets despite ranking third overall. It also leads four of the five speech-containing composition signatures evaluated in Appx.~\ref{app:capability}. Each subset evaluates the complete task requirements of instances containing that role. Their different leaders reveal complementary strengths that the overall ranking alone cannot express.

The disaggregated results show why a reference role must be interpreted in context. MiniMax-H3 leads subject+speech at \speechOnlyMini{} expected win rate, compared with \speechOnlySeed{} for Seedance-2.5. In subject+motion+speech, Seedance-2.5 leads at \speechMotionSeed{}, while MiniMax-H3 reaches \speechMotionMini{}. These signatures contain different instances; their contrast describes context-specific strengths and motivates controlled studies of how speech and motion requirements interact. The camera subset offers a different picture: the top three estimates cluster within one percentage point. The most informative outcome is therefore the pattern and size of the differences, rather than the number of columns won.

Seedance-2.0 ranks second overall and in every image-reference bucket, while MiniMax-H3 combines a speech advantage with lower estimates on motion-containing instances. Their relative order reverses between those two contexts. Overall strength and context-specific advantages therefore provide complementary guidance for comparing systems.

\subsection{Quality, Reference Affinity, and Task Fulfillment}
\begin{table}[t]
\centering
\definecolor{ORAVoutput}{HTML}{70EBAF}
\caption{\emph{Diagnostic measurements and ORAV performance.} Diagnostic means weight instances measurable for all five systems equally. Comp.~$z$: VBench-style quality; TQ: raw DOVER++ technical score $\times100$; AP: Aesthetic Predictor V2.5. Subject/Scene are full-frame DINO/CLIP similarities $\times100$; PQ is Audiobox production quality; Voice is ECAPA cosine. CER/WER are character/word error rates (\%); insertions can yield errors above 100\%. ORAV reports Davidson--Bradley--Terry expected win rates (\%) on shared-delivery instances. Kling-v3-Omni receives audio through a retained-soundtrack adapter because its tested interface lacks an audio-reference channel.}
\label{tab:fidelity}
\small
\setlength{\tabcolsep}{1.7pt}
\renewcommand{\arraystretch}{1.14}
\begin{tabular*}{\linewidth}{@{\extracolsep{\fill}}l*{10}{c}@{}}
\toprule
\multirow{2}{*}{System} &
\multicolumn{3}{c}{\cellcolor{ORAVvideo!14}Visual quality $\uparrow$} &
\multicolumn{2}{c}{\cellcolor{ORAVimage!14}Ref. affinity $\uparrow$} &
\multicolumn{1}{c}{\cellcolor{ORAVaudio!14}Audio $\uparrow$} &
\multicolumn{3}{c}{\cellcolor{ORAVaudio!14}Speech} &
\multicolumn{1}{c}{\cellcolor{ORAVoutput!16}ORAV $\uparrow$} \\
\cmidrule(lr){2-4}\cmidrule(lr){5-6}\cmidrule(lr){7-7}\cmidrule(lr){8-10}\cmidrule(lr){11-11}
& \cellcolor{ORAVvideo!8}Comp.~$z$ & \cellcolor{ORAVvideo!8}TQ & \cellcolor{ORAVvideo!8}AP & \cellcolor{ORAVimage!8}Subject & \cellcolor{ORAVimage!8}Scene & \cellcolor{ORAVaudio!8}PQ & \cellcolor{ORAVaudio!8}Voice $\uparrow$ & \cellcolor{ORAVaudio!8}CER $\downarrow$ & \cellcolor{ORAVaudio!8}WER $\downarrow$ & \cellcolor{ORAVoutput!8}Win rate \\
\midrule
Seedance-2.5 & $+0.10$ & $-1.78$ & $4.95$ & $40.61$ & $77.87$ & $6.99$ & $0.45$ & $5.51$ & $5.58$ & $\mathbf{68.38}$ \\
Seedance-2.0 & $+0.17$ & $-0.67$ & $4.99$ & $39.29$ & $79.01$ & $7.11$ & $0.48$ & $19.17$ & $16.23$ & $62.69$ \\
MiniMax-H3 & $+0.13$ & $-1.30$ & $4.60$ & $40.15$ & $75.49$ & $6.82$ & $0.59$ & $14.76$ & $18.09$ & $54.00$ \\
Wan3.0-Video & $-0.16$ & $-1.91$ & $5.05$ & $38.94$ & $77.09$ & $7.12$ & $0.39$ & $10.27$ & $112.48$ & $35.04$ \\
Kling-v3-Omni & $-0.19$ & $-3.97$ & $4.46$ & $40.45$ & $77.64$ & $6.72$ & $1.00$ & $187.84$ & $178.14$ & $29.90$ \\
\bottomrule
\end{tabular*}
\end{table}

The diagnostic results distinguish rendering an output, preserving source properties, and using those properties as instructed. Tab.~\ref{tab:fidelity} provides reproducible pointwise diagnostics of visual quality, reference affinity, audio production quality, and speech alongside ORAV performance. Each diagnostic uses shared measurable instances to explain differences in overall task performance.

Rendering quality and task fulfillment expose different strengths among the leading systems. Seedance-2.0 leads the visual composite and technical quality, Seedance-2.5 leads ORAV, and Wan3.0-Video has the highest mean aesthetic score. Comp.~$z$ and ORAV are positively associated across the five systems, with Spearman $\rho=0.70$; the quality ordering persists on ORAV's 350 shared-delivery instances. The paired bootstrap in Appx.~\ref{app:fidelity} more clearly supports Seedance-2.5's Comp.~$z$ advantage over Wan3.0-Video than the ordering among the leading three systems.

Reference affinity reveals why resemblance to a source is distinct from task success. Seedance-2.5 leads subject DINO similarity \citep{Caron_2021_ICCV}, with Kling-v3-Omni close behind despite ranking last on ORAV; Seedance-2.0 leads scene CLIP similarity \citep{pmlr-v139-radford21a}. These full-frame affinities measure resemblance to assigned sources, whereas Comp.~$z$ consistency measures stability within an output. Binding judgments assess whether the intended subject, motion, and scene are realized together in the requested event.

Speech diagnostics separate vocal resemblance from newly requested content. MiniMax-H3 has higher mean voice similarity than the other systems apart from Kling-v3-Omni, while Seedance-2.5 has the lowest mean CER and WER. Wan3.0-Video combines a high mean production-quality score with substantial word-error rates. Kling-v3-Omni's near-unit voice similarity reflects the retained-soundtrack adaptation used because its interface lacks a separate audio-reference channel.

Wan3.0-Video exhibits both recording replay and visual source carryover. The listener identifies reference-recording replay in \replayWan{} speech instances with available observations. Visual judgments also describe the motion source's performers and scene appearing in place of the instructed subjects and setting. These outputs retain reference information while changing the participants, setting, or utterance specified by the instruction.

These visual and auditory failures illustrate \emph{reference substitution}, in which unrequested source content replaces the intended event. Across the tested system configurations, source carryover or replay is mentioned in \substitutionShare{} of explanations for outputs judged unusable, as summarized in Appx.~\ref{app:failure}. These outcomes make selective reference use a concrete challenge: preserving the requested properties and binding them to the specified participants, actions, and scene.

These findings motivate instruction-conditioned reference representations and training examples that separate transferable properties from source-specific context. The same motion should support different performers and settings, and the same voice should support new utterances. Progress should be assessed by how reliably these properties are recomposed across reference combinations. ORAV links this goal to reference-level evidence and overall task preferences.

\section{Conclusion}
ORAV Bench makes multimodal context understanding and generation testable through semantic reference contracts. Its pairwise evaluator uses visual and auditory evidence to weigh differences in reference use by their impact on task fulfillment, achieving \humanAgreementSelected{} effective human agreement on held-out instances. Across five systems, strengths vary with reference composition, while reference substitution exposes a gap between source resemblance and task success. These findings motivate omni-reference generators that transfer assigned properties into new contexts, exclude conflicting source content, and bind the selected properties to their intended targets in the requested event.

\bibliography{refs}
\bibliographystyle{iclr2027_conference}

\appendix
\raggedbottom
\section{Supplementary Results and Evaluation Details}

\FloatBarrier
\subsection{Ranking, Coverage, and Exact Pairwise Records}
\label{app:ranking}
\begin{figure}[htbp]\centering
\includegraphics[width=\linewidth]{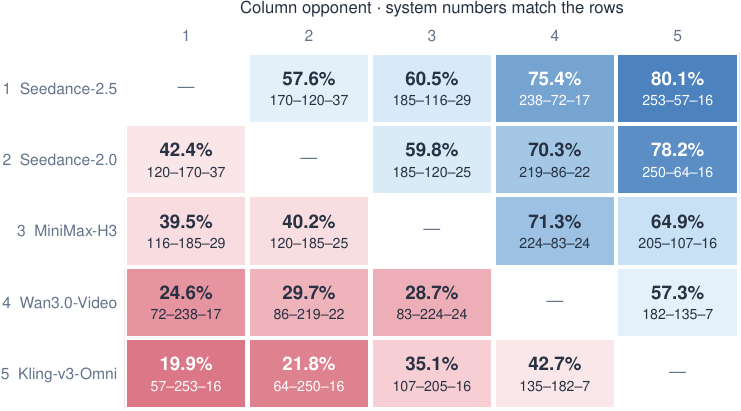}
\caption{\emph{Exact pairwise outcomes on shared-delivery instances.} Each cell gives the row system's empirical win rate, $(W+T/2)/(W+L+T)$, above its wins--losses--ties record. Ties include order conflicts; unavailable judgments are excluded. Column numbers identify the systems listed in the rows. Color encodes the win rate, with blue above and red below 50\%.}\label{fig:ranking_matrix}\label{tab:headtohead}
\end{figure}

The overall ranking separates task performance from delivery. Tab.~\ref{tab:ranking} reports delivery and evaluation coverage; Fig.~\ref{fig:ranking_matrix} preserves the exact pairwise records on the 350 shared-delivery instances. Each instance contributes at most one outcome per system pair after AB/BA fusion. Ranking ties include genuine ties and order conflicts; unavailable or evidence-insufficient comparisons are excluded.

For systems $i$ and $k$ in a given subset, let $w_{ik}$ count preferences for $i$ over $k$, and let $t_{ik}=t_{ki}$ count ranking ties. The Davidson extension of Bradley--Terry assigns each system a log-strength $\lambda_i$ and uses a common tie parameter $\nu>0$ within the fit \citep{bradley1952rank,davidson1970extending}:
\[
    D_{ik}=e^{\lambda_i}+e^{\lambda_k}
    +\nu e^{(\lambda_i+\lambda_k)/2},
\]
\[
    P(i\succ k)=\frac{e^{\lambda_i}}{D_{ik}},
    \qquad
    P(i\sim k)=\frac{\nu e^{(\lambda_i+\lambda_k)/2}}{D_{ik}}.
\]
Under $\sum_i\lambda_i=0$, we jointly maximize the log-likelihood over $\boldsymbol\lambda$ and $\nu$:
\[
    \ell(\boldsymbol\lambda,\nu)=\sum_{i<k}\left[
    w_{ik}\log P(i\succ k)+w_{ki}\log P(k\succ i)
    +t_{ik}\log P(i\sim k)\right].
\]
With $\widehat P$ denoting probabilities from the fitted model, the expected win rate against a uniformly selected opponent gives half credit to ties:
\[
    R_i=\frac{1}{K-1}\sum_{k\ne i}
    \left[\widehat P(i\succ k)+\tfrac12\widehat P(i\sim k)\right],
    \qquad K=5.
\]
Tab.~\ref{tab:overall} reports $\hat\lambda_i$ as overall strength; its role columns and the ORAV column of Tab.~\ref{tab:fidelity} report $100R_i$. Each role- or composition-conditioned fit re-estimates both the strengths and $\nu$ within its own subset. Fig.~\ref{fig:ranking_matrix} instead reports empirical win rates from the observed counts.

The order is unchanged across all \numVariants{} combinations of balanced versus all-instance scope, conflict-as-tie versus conflict exclusion, and Davidson, Rao--Kupper, or half-credit Bradley--Terry aggregation. A separate counterfactual that counts undelivered candidates as defeats also preserves the order. Ranking intervals use 2,000 bootstrap replicates resampling whole instances.

\begin{table}[htbp]\centering
\caption{\emph{Delivery and evaluation coverage.} Delivered videos are counted over all 380 task instances. Usable comparisons are the system's resolved or tied comparisons divided by attempted comparisons between delivered candidates, across all instances. These denominators differ from the 350-instance ranking subset. Neither missing videos nor unavailable judgments are counted as task defeats.}\label{tab:ranking}
\small\setlength{\tabcolsep}{4pt}
\begin{tabular*}{\linewidth}{@{\extracolsep{\fill}}lrrrr@{}}\toprule
System & Delivered & Rate & Usable comparisons & Rate \\
\midrule
Seedance-2.5 & 362/380 & 95.3\% & 1344/1436 & 93.6\% \\
Seedance-2.0 & 359/380 & 94.5\% & 1340/1427 & 93.9\% \\
MiniMax-H3 & 375/380 & 98.7\% & 1373/1459 & 94.1\% \\
Wan3.0-Video & 376/380 & 98.9\% & 1368/1464 & 93.4\% \\
Kling-v3-Omni & 380/380 & 100.0\% & 1375/1472 & 93.4\% \\
\bottomrule\end{tabular*}\end{table}

\FloatBarrier
\subsection{Performance by Reference Composition}
\label{app:capability}
Composition signatures reveal interactions that role-conditioned means can conceal. MiniMax-H3 leads four of the five displayed speech signatures, while Seedance-2.5 leads subject+motion+speech. These shifts show that role averages summarize complete compositions, whose co-occurring requirements can change the leading system.
\begin{figure}[htbp]\centering
\includegraphics[width=\linewidth]{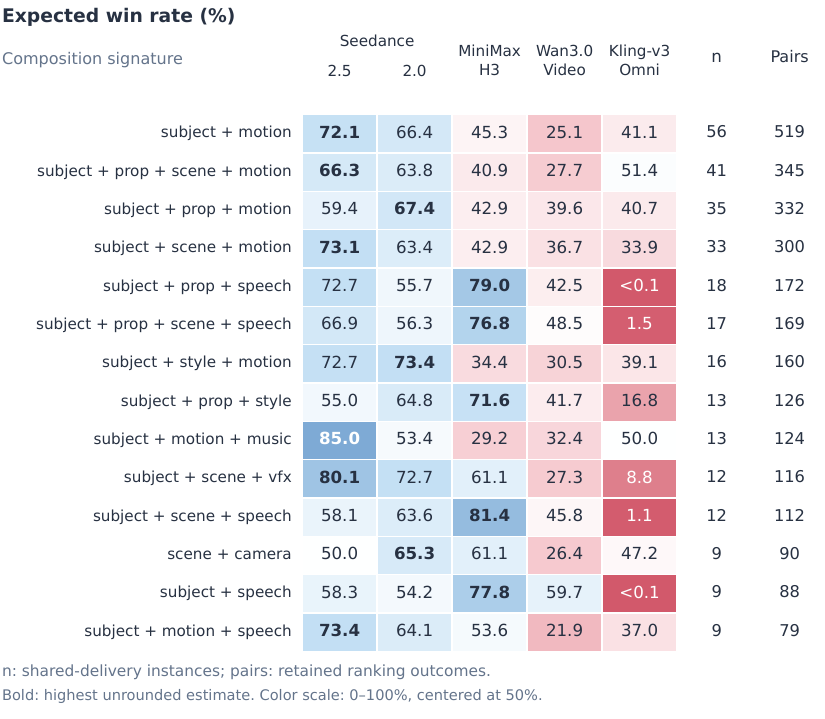}
\caption{\emph{Performance across composition signatures.} Expected win rates are refit within the 14 signatures with at least eight shared-delivery instances. Bold marks the highest unrounded estimate; $n$ gives shared instances and pairs gives retained ranking outcomes.}\label{fig:capability}
\end{figure}

\paragraph{Ranking without audio references.}
Kling-v3-Omni receives audio references through a retained-soundtrack adapter, while its tested interface directly accepts image and video references. We compare the five systems on tasks with entirely visual references. The 281 instances without speech or music references include 259 with outputs from all five systems. We refit Davidson--Bradley--Terry on these instances, retain order conflicts as ties, and judge both audio and video.

\begin{table}[htbp]\centering
\definecolor{ORAVoutput}{HTML}{70EBAF}
\caption{\emph{System ranking without audio references.} Davidson--Bradley--Terry estimates use 2,406 comparisons on 259 shared-delivery instances with no speech or music references. Intervals use 2,000 instance-bootstrap replicates. Strengths sum to zero; win rates (\%) are expected against a uniformly chosen other system. Bold marks the highest estimate.}\label{tab:no-audio-reference}
\small\setlength{\tabcolsep}{4pt}\renewcommand{\arraystretch}{1.14}
\begin{tabularx}{\linewidth}{@{}cl>{\centering\arraybackslash}Xc>{\centering\arraybackslash}X@{}}\toprule
\# & System & \cellcolor{ORAVoutput!16}Strength & \cellcolor{ORAVoutput!16}95\% CI & \cellcolor{ORAVoutput!16}Win rate (\%) \\\midrule
1 & Seedance-2.5 & $\mathbf{+0.67}$ & $[+0.51, +0.84]$ & $\mathbf{67.93}$ \\
2 & Seedance-2.0 & $+0.53$ & $[+0.39, +0.68]$ & $64.24$ \\
3 & MiniMax-H3 & $-0.03$ & $[-0.17, +0.10]$ & $49.18$ \\
4 & Kling-v3-Omni & $-0.53$ & $[-0.69, -0.37]$ & $35.76$ \\
5 & Wan3.0-Video & $-0.64$ & $[-0.84, -0.47]$ & $32.89$ \\
\bottomrule\end{tabularx}\end{table}

The top three systems retain their order from the full benchmark. MiniMax-H3's expected win rate decreases from 54.00\% to 49.18\%, consistent with its strong performance on speech-containing compositions. Kling-v3-Omni rises from fifth to fourth, ahead of Wan3.0-Video in 81.25\% of bootstrap fits. These rankings show how system strengths vary across reference compositions.

\FloatBarrier
\subsection{Diagnostic Measures and Aggregation}
\label{app:diagnostics}
Pointwise diagnostics distinguish output quality, reference affinity, and requested content. They operate on the same 1,852 delivered candidates used in ORAV. Each reported system mean gives equal weight to the instances on which all five systems have a valid value for that metric. Shared instance counts vary with the required reference role and observability, as recorded for the main comparison in Tab.~\ref{tab:diagnostic-support}. Missing audio, undetected faces, and invalid outputs are excluded rather than assigned a score of zero. Correlations use the same candidates for both measurements; their sample sizes count candidates, not independent instances.

For reference affinity, scores are averaged over sampled frames and references of the same role. A role-level value requires all its references to be measurable. Confidence intervals resample instances 2,000 times. Paired changes use identical instance--candidate pairs. These descriptive analyses do not fit human labels or correct for multiple comparisons. Correspondence with ORAV concerns the complete task, including its co-occurring roles.

The diagnostics use published pretrained models: CLIP/DINO for visual affinity, ECAPA and Whisper for voice and speech, and CoTracker3 for motion. ArcFace \citep{Deng_2019_CVPR} and ALADIN \citep{Ruta_2021_ICCV} provide face and style features. DOVER++, Aesthetic Predictor V2.5, Audiobox Aesthetics, NISQA, and DNSMOS predict output quality.

\begin{table}[htbp]
\centering
\definecolor{ORAVoutput}{HTML}{70EBAF}
\caption{\emph{Shared instance counts for the main comparison.} Diagnostic counts require valid values from all five systems for each metric; Subject uses DINO and Scene uses CLIP. ORAV uses instances with all five outputs delivered, retaining 3,280 available pairwise outcomes for the strength fit.}\label{tab:diagnostic-support}
\small\setlength{\tabcolsep}{1.7pt}
\renewcommand{\arraystretch}{1.14}
\begin{tabular*}{\linewidth}{@{\extracolsep{\fill}}l*{10}{r}@{}}
\toprule
Measure & \cellcolor{ORAVvideo!14}Comp.~$z$ & \cellcolor{ORAVvideo!14}TQ & \cellcolor{ORAVvideo!14}AP & \cellcolor{ORAVimage!14}Subject & \cellcolor{ORAVimage!14}Scene & \cellcolor{ORAVaudio!14}PQ & \cellcolor{ORAVaudio!14}Voice & \cellcolor{ORAVaudio!14}CER & \cellcolor{ORAVaudio!14}WER & \cellcolor{ORAVoutput!14}ORAV \\
\midrule
Shared instances ($n$) & 350 & 350 & 350 & 332 & 166 & 91 & 72 & 52 & 22 & 350 \\
\bottomrule
\end{tabular*}
\end{table}

\begin{table}[htbp]\centering
\caption{\emph{Pretrained components for affinity, speech, and motion diagnostics.} Component names identify the evaluated variants; references identify their published methods.}\label{tab:diagnostic-sources}
\small\setlength{\tabcolsep}{3pt}
\begin{tabular*}{\linewidth}{@{\extracolsep{\fill}}lll@{}}\toprule
Measurement & Component & Source \\\midrule
Visual affinity & CLIP ViT-B/32 & \citet{pmlr-v139-radford21a} \\
Visual affinity & DINOv1 ViT-B/16 & \citet{Caron_2021_ICCV} \\
Voice similarity & ECAPA-TDNN & \citet{desplanques20_interspeech} \\
Speech text & Whisper-small & \citet{pmlr-v202-radford23a} \\
Motion direction & CoTracker3 scaled\_offline & \citet{Karaev_2025_ICCV} \\
\bottomrule\end{tabular*}
\end{table}

\FloatBarrier
\subsection{Visual Quality and Reference Fidelity}
\label{app:fidelity}
The VBench-style control \citep{huang2024vbench} uses DINO for subject consistency, CLIP for background consistency, AMT-S \citep{Li_2023_CVPR} for motion smoothness, CLIP/LAION for aesthetics, and MUSIQ-SPAQ \citep{9710973} for imaging quality. Each dimension is standardized over all 1,852 candidates, and their unweighted mean gives Comp.~$z$. It is a corpus-relative control, not a published VBench leaderboard score. The quality and ORAV rankings are positively associated across five systems, with Spearman $\rho=0.70$, while differing among the leaders. Restricting these corpus-standardized scores to 350 shared-delivery instances preserves the quality order. Paired intervals include zero for adjacent leading systems; the Seedance-2.5--Wan3.0-Video gap is \qualityWideGap{}, with a 95\% confidence interval \qualityWideCI{}.

\begin{figure}[htbp]\centering
\includegraphics[width=\linewidth]{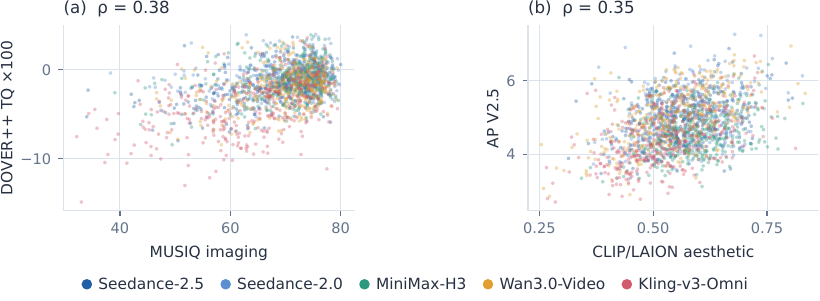}
\caption{\emph{Quality models capture different aspects of the same outputs.} Each panel compares two measures on the same 1,852 candidates, colored by system. DOVER++ technical quality and AP V2.5 aesthetics show partial correspondence with MUSIQ imaging quality and CLIP/LAION aesthetics, respectively. Spearman correlations summarize the overlap between these quality measures.}\label{fig:diag-quality}
\end{figure}

The additional quality models measure related but distinct properties, as Fig.~\ref{fig:diag-quality} shows. DOVER++ \citep{Wu_2023_ICCV} uses official fragment sampling: three 32-frame technical clips and a 32-frame aesthetic branch, with technical and aesthetic scores reported separately. Aesthetic Predictor V2.5 averages eight temporal midpoint frames using its SigLIP-SO400M processor \citep{Zhai_2023_ICCV}. On 350 common instances, Seedance-2.5 minus Seedance-2.0 has a raw technical-quality difference of $-0.0111$ with 95\% interval $[-0.0128,-0.0095]$. Wan3.0-Video has the highest V2.5 mean, but its difference from Seedance-2.0 is $0.057$ with interval $[-0.010,0.123]$.

\begin{table}[htbp]
\centering
\caption{\emph{Additional visual-quality dimensions.} VBench-style consistency, smoothness, and aesthetic scores are multiplied by 100; imaging quality retains its 0--100 scale. DOVER++ AQ reports its raw aesthetic score. Means use the shared instances in each column. Higher values indicate better quality within each measure; scales differ across columns.}\label{tab:diag-quality}
\small\setlength{\tabcolsep}{3pt}
\renewcommand{\arraystretch}{1.14}
\begin{tabular*}{\linewidth}{@{\extracolsep{\fill}}lrrrrrr@{}}
\toprule
System & Subj. & Bkgd. & Smooth. & Aesth. & Imag. & AQ \\
Shared instances ($n$) & 350 & 350 & 350 & 350 & 350 & 350 \\
\midrule
Seedance-2.5 & 91.6 & 93.9 & 99.3 & 56.3 & 67.3 & +0.035 \\
Seedance-2.0 & 92.4 & 93.3 & 99.2 & 57.0 & 71.0 & +0.040 \\
MiniMax-H3 & 90.6 & 92.2 & 99.3 & 59.4 & 70.7 & +0.036 \\
Wan3.0-Video & 89.8 & 91.2 & 99.1 & 54.7 & 70.0 & +0.038 \\
Kling-v3-Omni & 90.9 & 93.0 & 99.1 & 53.1 & 64.2 & +0.028 \\
\bottomrule
\end{tabular*}
\end{table}

Reference similarity addresses a different question: which source properties are visible in the output? CLIP and DINO compare eight uniformly spaced midpoint frames with the assigned image or sampled video references, averaging frame-pair cosine similarities. Full-frame inputs preserve aspect ratio and are padded to 224 pixels. Localized affinity uses GroundingDINO \citep{10.1007/978-3-031-72970-6_3} boxes with within-category one-to-one matching; scene comparisons use SAM2.1 \citep{ravi2025sam} foreground masks to contrast the assigned scene with the motion-source background.

On identical non-tied pairs with equal instance weights, neither localization nor target-minus-source affinity shows a clear gain in correspondence with ORAV preferences over its full-frame or target-only counterpart; all paired 95\% intervals include zero.

Face and style measurements provide feature-specific views in Tab.~\ref{tab:diag-reference}. ArcFace uses detected single-face references, grayscale bounding-box crops, and concatenated original/flip features without landmark alignment. Its correlation with localized DINO is $0.256$ over 1,542 joint candidates, showing limited correspondence between face identity and whole-subject appearance.
\begin{table}[htbp]
\centering
\caption{\emph{Complementary reference diagnostics.} Face cosine uses ArcFace on detected faces; style cosines use ALADIN, CLIP, and DINO; motion compares visible foreground track directions. Each column uses its own five-system common subset. These affinities describe the measured feature and do not establish the identity of the actor performing an action or the correctness of the complete task.}\label{tab:diag-reference}
\small\setlength{\tabcolsep}{3pt}
\renewcommand{\arraystretch}{1.14}
\begin{tabular*}{\linewidth}{@{\extracolsep{\fill}}lrrrrr@{}}
\toprule
System & Face & ALADIN & CLIP-Style & DINO-Style & Motion dir. \\
Shared instances ($n$) & 252 & 43 & 43 & 43 & 73 \\
\midrule
Seedance-2.5 & 0.360 & 0.324 & 0.627 & 0.272 & 0.290 \\
Seedance-2.0 & 0.347 & 0.349 & 0.621 & 0.286 & 0.301 \\
MiniMax-H3 & 0.336 & 0.246 & 0.610 & 0.244 & 0.311 \\
Wan3.0-Video & 0.332 & 0.371 & 0.643 & 0.342 & 0.572 \\
Kling-v3-Omni & 0.234 & 0.287 & 0.635 & 0.309 & 0.606 \\
\bottomrule
\end{tabular*}
\end{table}

CoTracker3 compares visible moving-track directions on 32 relative-time frames, matching foreground tracks in both directions. The common foreground subset of 73 instances supports inspection of motion orientation. This score measures direction under relative-time alignment; it does not establish speed fidelity or the identity of the acting subject.

\FloatBarrier
\subsection{Auditory Quality and Selective Reference Use}
\label{app:audio-diagnostics}
Speech measurements separate listening quality, vocal identity, and the specified utterance. Audiobox production quality \citep{11434623} averages ten-second windows by duration. ECAPA-TDNN compares VAD-selected speech with the voice reference. Whisper-small transcription is normalized for punctuation and case, with character error rate for targets containing Chinese characters and word error rate for other targets. Normalized edit distance includes insertions and can exceed 100\%. Recording reuse is measured by the maximum normalized waveform correlation over time shifts at 4\,kHz, requiring at least three seconds and half the shorter recording to overlap.

These measurements reveal complementary differences. MiniMax-H3 has higher voice affinity than Seedance-2.5, Seedance-2.0, and Wan3.0-Video, while Seedance-2.5 has lower mean errors on the requested text. Wan3.0-Video's production-quality mean is 7.12 even though its word-error rate is 112.5\%; the listener also identifies recording replay in \replayWan{} assessed speech outputs.

For Kling-v3-Omni, the tested interface lacks an independent audio-reference channel, so the retained-soundtrack adapter provides audio through video input. Its voice cosine of 0.998 and waveform correlation of 0.9998 quantify recording retention under this configuration.

\begin{figure}[htbp]\centering
\includegraphics[width=\linewidth]{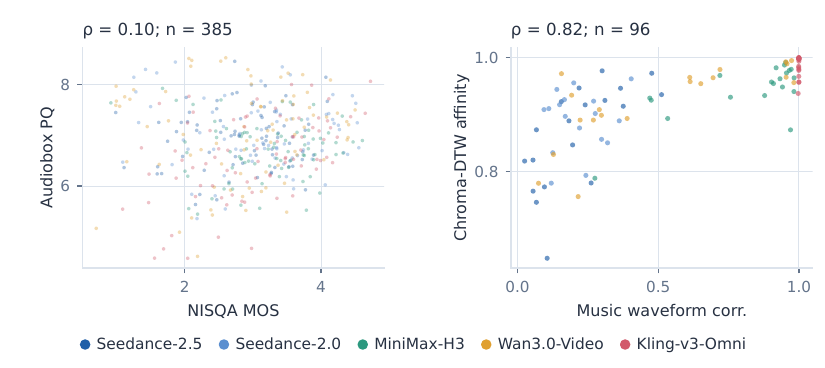}
\caption{\emph{Auditory quality and music-reference use.} Left: Audiobox PQ versus NISQA MOS on 385 speech candidates. Right: chroma-DTW affinity versus waveform reuse on 96 music candidates. High music affinity can reflect retained source audio.}\label{fig:diag-complementary}
\end{figure}
\begin{table}[htbp]
\centering
\caption{\emph{Speech quality and music-reference affinity.} NISQA MOS and DNSMOS OVRL measure speech quality. CLAP, MuQ, chroma-DTW, and waveform correlation compare music references with the output. Music affinity has no universal quality direction: a task may request a new composition rather than the source recording. Kling-v3-Omni uses retained reference soundtracks.}\label{tab:diag-audio}
\small\setlength{\tabcolsep}{3pt}
\renewcommand{\arraystretch}{1.14}
\begin{tabular*}{\linewidth}{@{\extracolsep{\fill}}lrrrrrr@{}}
\toprule
System & NISQA & DNSMOS & CLAP & MuQ & Chroma & Waveform \\
Shared instances ($n$) & 74 & 72 & 17 & 17 & 17 & 17 \\
\midrule
Seedance-2.5 & 2.83 & 2.80 & 0.404 & 0.668 & 0.884 & 0.229 \\
Seedance-2.0 & 3.11 & 2.97 & 0.509 & 0.763 & 0.890 & 0.216 \\
MiniMax-H3 & 3.20 & 2.90 & 0.562 & 0.783 & 0.940 & 0.800 \\
Wan3.0-Video & 2.89 & 2.76 & 0.560 & 0.792 & 0.925 & 0.508 \\
Kling-v3-Omni & 3.01 & 2.68 & 0.661 & 0.929 & 0.988 & 1.000 \\
\bottomrule
\end{tabular*}
\end{table}

Audiobox PQ and NISQA MOS correlate only weakly on the same 385 speech candidates, with $\rho=0.103$; NISQA and DNSMOS correlate more strongly, with $\rho=0.578$ on 381 candidates. These models target production quality, speech impairment \citep{mittag21_interspeech}, and denoising quality \citep{9746108}, respectively, providing complementary views of the generated soundtrack.

Music diagnostics use the mixed output audio. CLAP \citep{10095969} and MuQ \citep{11134793} provide pretrained embedding affinity; chroma-DTW measures pitch-class sequence resemblance. Across 96 candidates, chroma affinity and waveform reuse correlate at $\rho=0.825$. The five-system common music subset contains 17 instances, including different instructions for retaining or recreating musical content. These scores expose reference use without a universal upward ranking.

\FloatBarrier
\subsection{Reference Substitution and Task Fulfillment}
\label{app:failure}
Reference-use failures distinguish source resemblance from task fulfillment. Tab.~\ref{tab:failure} separates candidate-level usability judgments, substitution cues, and unique-output replay counts. Kling-v3-Omni's replay count characterizes the retained-soundtrack adapter used by its tested interface.

\begin{table}[H]
\centering
\caption{\emph{Reference-use failures.} Unusability rates count ordered candidate judgments; substitution cues are a share of vetoes. Replay counts use unique outputs for speech instances.}
\label{tab:failure}
\small
\setlength{\tabcolsep}{3pt}
\renewcommand{\arraystretch}{1.14}
\begin{tabular*}{\linewidth}{@{\extracolsep{\fill}}lrrrrr@{}}
\toprule
\multirow{2}{*}{System} & \multicolumn{3}{c}{Ruled unusable} & Substitution cues & Reference voice \\
\cmidrule(lr){2-4}
& All & Video ref. & No video ref. & Share of vetoes & Replayed \\
\midrule
Seedance-2.5 & 2.3\% & 3.2\% & 0.0\% & 95\% & 0/75 \\
Seedance-2.0 & 4.7\% & 6.0\% & 1.4\% & 91\% & 0/75 \\
MiniMax-H3 & 7.7\% & 10.7\% & 0.0\% & 94\% & 0/78 \\
Wan3.0-Video & 43.8\% & 55.9\% & 12.0\% & 94\% & 9/78 \\
Kling-v3-Omni & 32.0\% & 31.4\% & 33.9\% & 96\% & 79/79 \\
\bottomrule
\end{tabular*}
\end{table}

Tab.~\ref{tab:failure} summarizes whether each candidate supplies a usable rendition and why. Candidates are vetoed in \vetoShare{} of candidate-level judgments. Vetoes determine \vetoDecided{} of resolved pairs, rising to \vetoDecidedVideo{} on instances with a video reference. In \substitutionShare{} of veto explanations, the text contains cues of \emph{reference substitution}: source content appearing in place of the requested result. Examples include transferring the motion reference's original performers and environment, or replaying a voice reference instead of producing the specified utterance. These are failures to use context as instructed, even when the output resembles a reference closely. The distinction is central to evaluating context use: the desired attributes must be composed into the requested event.

Wan3.0-Video's veto rate is \vetoWorstVideo{} with a video reference and \vetoWorstNoVideo{} without one, reflecting frequent reproduction of source scenes and performers. The listening stage also identifies reference-recording replay in \replayWan{} speech instances with available observations; no replay is observed for Seedance-2.5, Seedance-2.0, or MiniMax-H3. Its visual source carryover and recording replay illustrate how recognizable source information can displace the requested composition across modalities.

Tab.~\ref{tab:failure} counts candidate judgments across all successful ordered comparisons, including different opponents and both presentation orders. Unavailable responses are excluded. Substitution shares use a fixed vocabulary of source-carryover and replay terms in veto explanations, summarizing textual cues rather than independently annotated causes. Speech replay is counted once per instance and system, using the listening-stage observations.

\begin{figure}[htbp]
\centering
\includegraphics[width=\linewidth]{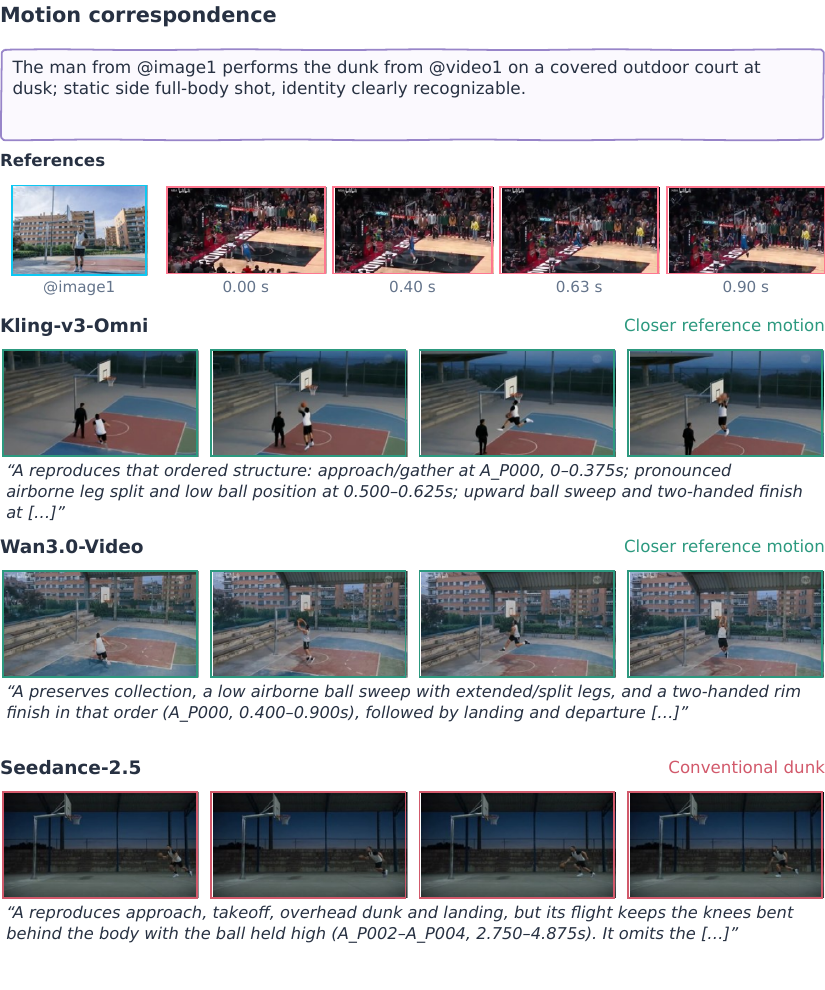}
\caption{\emph{Reference-specific motion beyond overall appearance.} In \texttt{subject-motion-7f681a}, forward leg extension with a lowered ball distinguishes the demonstrated dunk from a conventional rendition. Frames and verbatim quotations come from the recorded evaluation.}
\label{fig:cases-motion}\label{fig:cases}
\end{figure}

\begin{figure}[htbp]
\centering
\includegraphics[width=\linewidth]{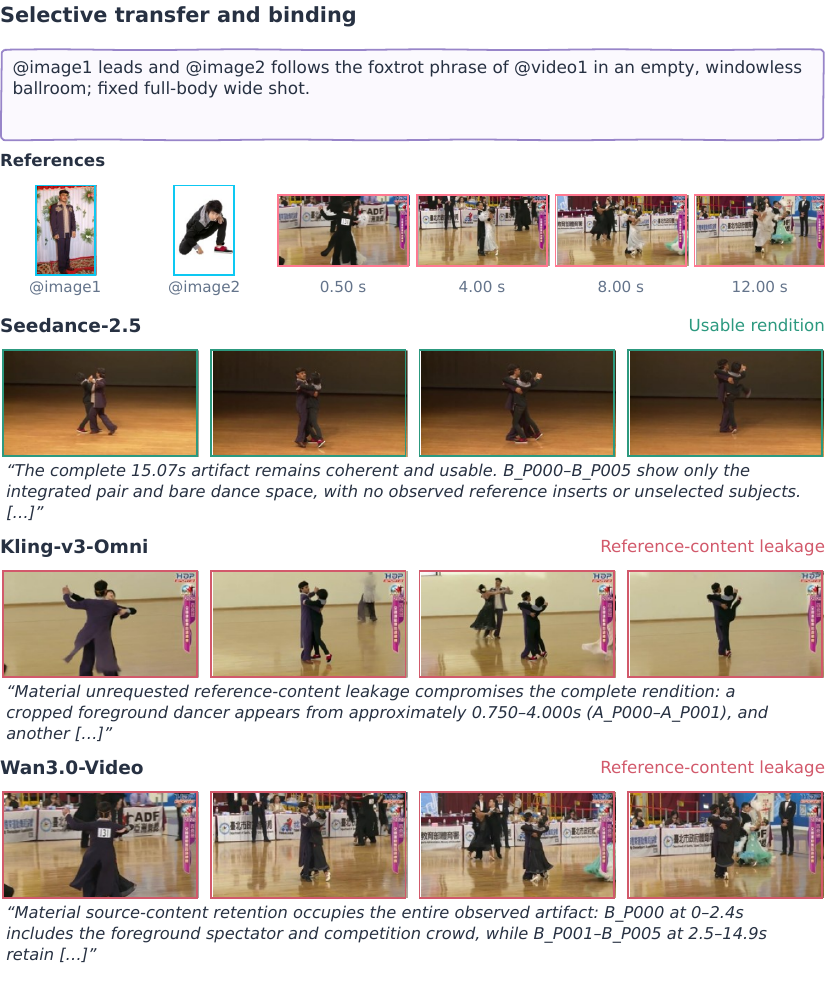}
\caption{\emph{Selective transfer of motion and performers.} In \texttt{subject-motion-be4824}, reproducing the competition hall and its original dancers replaces the requested performers and scene. Frames and verbatim quotations reveal reference substitution rather than task fulfillment.}
\label{fig:cases-binding}
\end{figure}

The recorded cases in Figs.~\ref{fig:cases-motion} and \ref{fig:cases-binding} illustrate how reference-specific evidence distinguishes the demonstrated movement and its intended performers from visually similar source content.
\label{app:cases}

\FloatBarrier
\subsection{Evaluation Scope}
\label{app:limits}
ORAV assesses multimodal context understanding and generation through the complete system's outputs. Reference-level judgments explain their relation to the task; they do not isolate the model's internal understanding from synthesis. Timestamped panels and auditory observations support inspection of the evidence, while fine-grained audio-video synchronization remains outside the validated measurement scope. The human-validation subset was held out from evaluator development, and its labels are used only for independent validation.

\FloatBarrier
\subsection{Candidate Provenance}
\label{app:provenance}
Of 1,900 system--instance slots, 1,852 contributed a video and 48 did not. Content policies and related service restrictions account for undelivered videos and unavailable judge responses. Each delivered slot contributes one candidate. The manifest records instance and media identities, delivery status, and tested system configurations.

Each ordered comparison contributes at most one valid response. Failed requests may be retried with unchanged inputs and settings, without selection among valid responses.

Kling-v3-Omni's tested interface accepts image and video references but no independent audio input. We package each audio reference as a video-input soundtrack, using the task's video reference when available and a static reference image otherwise. Speech instances use original-sound retention. This compatibility adapter lets the system receive the task's audio content through its supported interface.

\FloatBarrier
\subsection{Evaluation Settings and Human Validation}
\label{app:prompt}
\begin{table}[htbp]\centering
\caption{\emph{Judge and listener settings.} Output budgets are in tokens.}\label{tab:evaluation-settings}
\small\setlength{\tabcolsep}{3pt}
\begin{tabular*}{\linewidth}{@{\extracolsep{\fill}}lrrr@{}}\toprule
Component & Reasoning & Temperature & Output budget \\\midrule
GPT-6-Astra judge & medium & --- & 8,192 \\
Gemini-3.1-Pro listener & medium & 0.3 & 32,768 \\
Gemini-3.1-Pro vanilla & high & 0.3 & 16,384 \\
\bottomrule\end{tabular*}
\par\smallskip
\noindent Reasoning denotes effort or thinking level; --- denotes default temperature without an explicit override.
\end{table}

The reasoning judge receives the English instruction, visual references, chronological candidate panels, and textual auditory observations. Frames are sampled at 8 fps, resized to width 512 with aspect ratio preserved, and arranged in at most six panels per clip, bounded by 2048 pixels per side. Row-major grids retain all frames at the largest fitting scale. Both Gemini components use Gemini-3.1-Pro-Preview. Vanilla receives native videos and references with default sampling; its prompt requests reference fidelity and audio-video quality in both presentation orders.

Each valid response provides reference-level, binding, instruction, usability, and overall preferences. Contract targets, requirements, and exclusions are interpreted within the request, not supplied as separate annotations. Ranking compares identity-mapped overall preferences across orders. The implementation retains the complete prompts and response schemas.

Human agreement averages eligible reviewer--pair agreement within each of the 38 held-out instances, then averages instances equally. Genuine ties and unavailable or order-conflicting automated decisions are excluded. Human rankings are not replaced with majority-vote labels. On common support, the W/T/L visualization retains genuine ties and counts automated order conflicts in its tie segment.

\paragraph{Validating the ORAV judge.}
ORAV evaluation connects observations of individual references to the success of the requested event. Timestamped panels expose motion phases, and the listener distinguishes voice characteristics, spoken content, and recording reuse. The decision protocol then weighs these observations according to their consequences for task fulfillment. We examine this design through two paired comparisons with human rankings. Simplified Astra retains ORAV's prepared evidence and uses a simple pairwise prompt to return an overall winner and a brief reason. Every pair is evaluated in both orders.

\begin{table}[htbp]\centering
\definecolor{ORAVoutput}{HTML}{70EBAF}
\caption{\emph{Paired human agreement on the 38 held-out instances.} Each row uses its own shared eligible comparisons. Scores follow evaluator order; $\Delta$ is their paired difference. Instances receive equal weight; 95\% intervals for $\Delta$ use 2,000 instance-bootstrap replicates.}\label{tab:paired-judge-validation}
\small\setlength{\tabcolsep}{4pt}\renewcommand{\arraystretch}{1.14}
\begin{tabularx}{\linewidth}{@{}l>{\centering\arraybackslash}X>{\centering\arraybackslash}X>{\centering\arraybackslash}X@{}}\toprule
Comparison & \cellcolor{ORAVoutput!16}Agreement (\%) & \cellcolor{ORAVoutput!16}$\Delta$ (pp) & \cellcolor{ORAVoutput!16}95\% CI ($\Delta$) \\\midrule
ORAV Judge vs. Simplified Astra & $86.97\;/\;79.52$ & $+7.45$ & $[+1.08, +15.83]$ \\
Simplified Astra vs. Vanilla Gemini & $82.09\;/\;71.89$ & $+10.20$ & $[+4.48, +17.74]$ \\
\bottomrule\end{tabularx}
\end{table}

\FloatBarrier

\emph{ORAV Judge vs.\ Simplified Astra.} Both settings use the same Astra model, visual and auditory evidence, reasoning-effort setting, and maximum output budget. The full protocol relates per-reference matches to binding and instruction fulfillment, then resolves competing local advantages using usability, task fulfillment, and general quality as ordered priorities. The 7.45-percentage-point gain supports ORAV's structured decision protocol over the simplified pairwise prompt under the same judge model and prepared evidence.

\emph{Simplified Astra vs.\ Vanilla Gemini.} Both use the same simple instruction to assess reference fidelity and audio-video quality. Vanilla Gemini reads native videos and references, while Simplified Astra combines the Astra judge with timestamped panels and listener observations. Human agreement is 10.20 points higher for the latter configuration. Together, these comparisons support preparing multimodal evidence and judging how reference contributions jointly fulfill the instruction.

\end{document}